\documentclass{article}

\usepackage[preprint]{neurips_2025}

\usepackage[utf8]{inputenc} 
\usepackage[T1]{fontenc}    
\usepackage{hyperref}       
\usepackage{url}            
\usepackage{booktabs}       
\usepackage{amsfonts}       
\usepackage{nicefrac}       
\usepackage{microtype}      
\usepackage{amsmath}
\usepackage{enumitem}
\usepackage{algorithm}
\usepackage{algpseudocode}
\usepackage{graphicx}
\usepackage{booktabs}       
\usepackage{multirow}       
\usepackage{colortbl}       
\usepackage[table]{xcolor}  
\usepackage{float}
\usepackage[most]{tcolorbox}
\usepackage{lmodern}
\usepackage{titlesec}

\titleformat{\section}{\bfseries\Large}{\thesection}{1em}{}

\usepackage{siunitx}        
\title{A*-Decoding: Token-Efficient Inference Scaling}

\author{%
  Giannis Chatziveroglou\thanks{Work done during time at Cohere.}\\
  MIT\\
  \texttt{gchatz@mit.edu} \\
}

\begin{document}

\maketitle

\begin{abstract}
  Inference-time scaling has emerged as a powerful alternative to parameter scaling for improving language model performance on complex reasoning tasks. While existing methods have shown strong performance gains under fixed compute budgets, there has been little focus on optimally utilizing that budget during inference. In this work, we introduce A*-decoding, a search-based inference-time strategy that builds on the A* search algorithm to optimally utilize a fixed compute budget by prioritizing high-quality reasoning paths during generation. We frame language model decoding as a structured search in a state space of partial solutions, applying the A* transition model to identify promising continuations guided by an external process supervision signal. In our experiments, A*-decoding reaches the performance levels of strong inference scaling baselines like best-of-N and particle filtering while using up to 3x fewer tokens and 30\% fewer PRM passes under equivalent compute budgets. On the MATH500 and AIME 2024 benchmarks, A*-decoding enables Llama-3.2-1B-Instruct to match the performance of the 70x larger Llama-3.1-70B-Instruct, and allows Qwen3-1.7B to reach o1-like reasoning accuracy. These results highlight the power of structured search in decoding, offering an alternative to brute-force sampling or scale-driven gains. Our work demonstrates how thoughtful inference-time strategies can enhance reasoning in SLMs, pointing toward future advances in more efficient and scalable language model deployment.

\end{abstract}

\begin{figure}
  \centering
  \includegraphics[width=.99\linewidth]{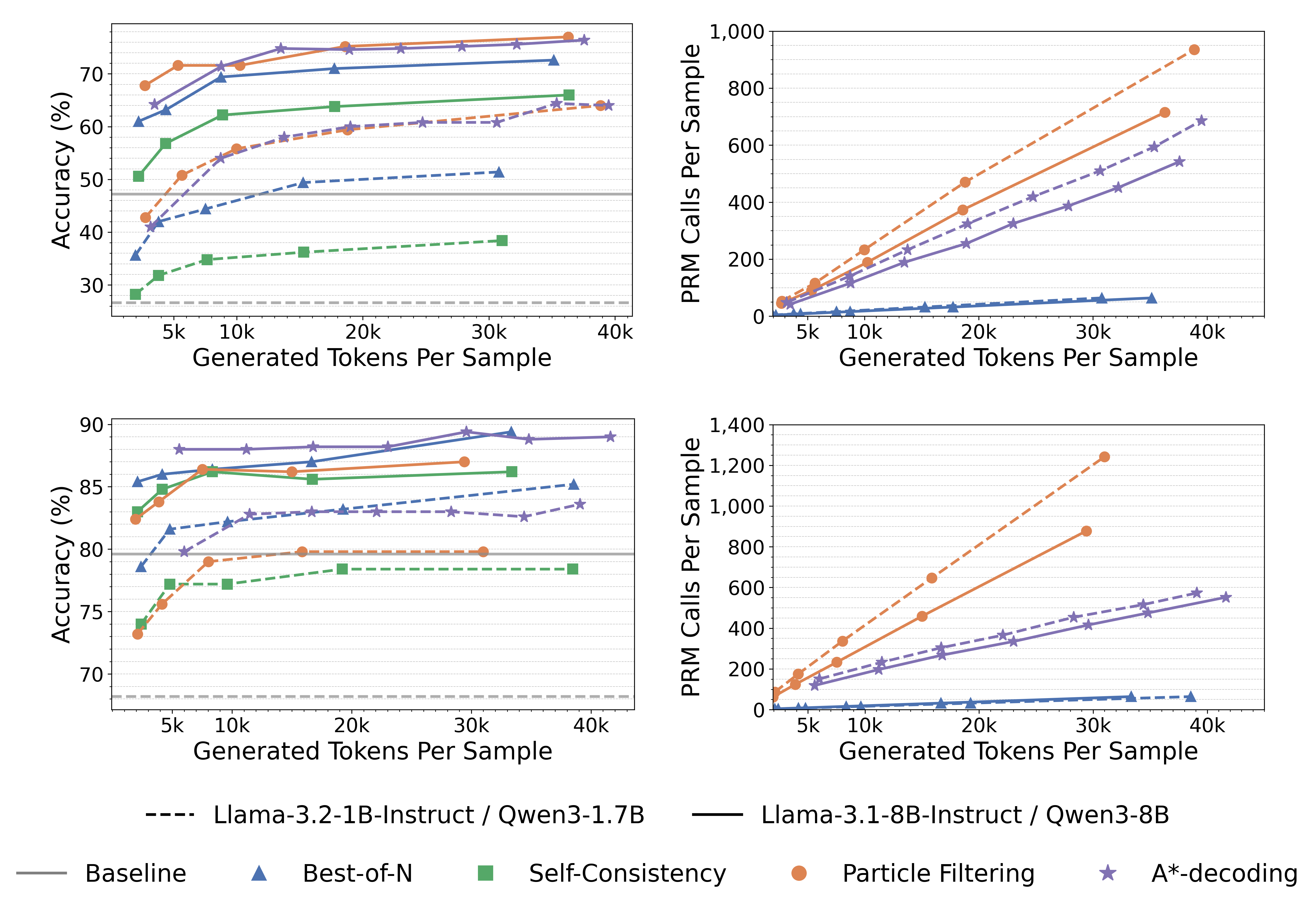}
  \caption{Top: Results with Llama models. Bottom: Results with Qwen models. Left: Token cost-performance frontier on MATH500. A*-decoding (ours) consistently achieves higher or equal accuracy compared to alternatives with lower token usage. Right: PRM cost-performance frontier on MATH500. A*-decoding optimally allocates PRM inference budget by focusing compute on promising partial trajectories. Points show results for 4–64 sampled generations; A*-decoding uses up to 16 for Llama, and 32 for Qwen.}
  \label{fig:cost_ablation}
\end{figure}

\section{Introduction}
Large language models (LLMs) have historically achieved improved performance primarily by increasing model size and training data. Empirical scaling laws show predictable loss reductions with increases in model size, dataset size, and training FLOPs \cite{kaplan2020scalinglawsneurallanguage}, driving the development of trillion-parameter models \cite{meta2025llama4,deepseekai2025deepseekr1incentivizingreasoningcapability,openai2025gpt45}. However, training and deploying such models requires massive specialized hardware, introducing significant capital and operational complexity. In response, recent work has shifted toward reallocating training compute to inference-time reasoning \cite{beeching2024scalingtesttimecompute}. Chain-of-thought (CoT) prompting \cite{wei2023chainofthoughtpromptingelicitsreasoning} spurred structured approaches like tree-of-thoughts and self-consistency \cite{yao2023treethoughtsdeliberateproblem, wang2023selfconsistencyimproveschainthought}, and models like OpenAI’s o1 show how computational effort at inference can boost reasoning \cite{openai2024o1}. This shift reframes inference as an active reasoning process, closer to search and planning than to passive next‑token prediction, enabling algorithms that strategically allocate compute.\\\\
Prior work on inference-time strategies has shown that small language models (SLMs) can rival larger models in reasoning accuracy, but such methods often overlook token-level efficiency. Snell et al. \cite{snell2024scalingllmtesttimecompute} demonstrate that SLMs can outperform significantly larger counterparts when inference-time compute is allocated optimally under fixed FLOPs budgets. While existing research highlights the potential of inference-time strategies to improve predictive accuracy \cite{shinn2023reflexionlanguageagentsverbal, yao2023reactsynergizingreasoningacting, huang2025bestofnbestthemcoverage, puri2025probabilisticinferenceapproachinferencetime, zhou2024languageagenttreesearch, guan2025rstarmathsmallllmsmaster, zhang2023planninglargelanguagemodels}, they primarily optimize for correctness, often by increasing the number or diversity of samples. In contrast, token efficiency, defined in terms of utility per generated token, has received comparatively limited attention in the literature, despite being a critical dimension of scaling with significant implications for latency and cost, particularly when deploying SLMs in resource-constrained settings. In this work, we focus on inference-time strategies that explicitly optimize the tradeoff between efficiency and performance, aiming to close the gap between small and large models through strategic computation rather than increased scale.\\\\
To address this gap, we propose A*-decoding. We posit that reframing autoregressive decoding as a state-space traversal guided by the A* search algorithm can deliver similar exact‑match accuracy as larger CoT models and inference-time scaling methods, while generating substantially fewer tokens under the same budget. This hypothesis is motivated by the algorithmic design of A* as a best-first, optimal shortest-path algorithm, which enables it to prioritize the most promising partial generations while efficiently pruning costly, less relevant candidates. Grounded in this hypothesis, we design a decoding framework that operationalizes A* search over a dynamically constructed graph of candidate next thoughts. At each generation step, our method samples a fixed number of candidate continuations from the policy, evaluates each using a composite score that combines a process-supervised heuristic with an inferred progress-based cost term, and expands the most promising state based the A* transition model until a complete output sequence is generated. To validate this framework, we conduct experiments on the MATH500 and AIME 2024 benchmarks using a range of SLMs including Llama3 and Qwen3 variants between 1-8B parameters. We evaluate A*-decoding against strong inference-time baselines such as best-of-N, self-consistency, and particle filtering. Our main contributions are:
\begin{enumerate}
\item \textbf{A novel framework for autoregressive generation.} We propose A*-decoding, a search-based decoding strategy that reimagines autoregressive inference as a best-first graph traversal, combining a learned heuristic and cost estimate to guide generation toward high-quality outputs with minimal token usage.
\item \textbf{Improved token efficiency without accuracy loss.} We demonstrate that A*-decoding achieves competitive exact-match accuracy compared to established inference-time baselines, including best-of-N, self-consistency, and particle filtering, while generating up to 3x fewer overall tokens and 30\% fewer inference passes.
\item \textbf{Strong empirical results on math reasoning.} Through extensive experiments on the MATH500 and AIME 2024, we show that SLMs (1–8B parameters) using A*-decoding can match or exceed the performance of much larger CoT-based models (up to 70B) and thinking models like OpenAI's o1.
\end{enumerate}
\section{Related Work}
\textbf{Decoding strategies} in language modeling have largely focused on maximizing likelihood during inference. While greedy decoding offers efficiency, its lack of exploration makes it vulnerable to early errors. Sampling-based methods, such as temperature and nucleus sampling, promote diversity by drawing from the full token distribution, though often at the cost of coherence and quality \cite{holtzman2020curiouscaseneuraltext, Renze_2024}. Beam search introduces parallel hypotheses to improve over greedy decoding but frequently converges on high-probability yet suboptimal sequences \cite{xie2023selfevaluationguidedbeamsearch, pmlr-v97-cohen19a}. More recently, self-consistency decoding has improved reliability by aggregating answers across multiple CoT rollouts, treating each as an independent sample \cite{wang2023selfconsistencyimproveschainthought}. However, these methods all generate text as flat sequences, lacking the ability to evaluate intermediate reasoning steps thus limiting their capacity to discard faulty paths or reuse promising ones. This motivates more strategic, inference-time approaches that reason over partial trajectories.
\\\\
\textbf{Process supervision} promotes logical correctness by evaluating intermediate reasoning steps, guiding models toward more coherent outputs. Process Reward Models (PRMs) implement this principle through lightweight classifiers trained on step-labeled traces to provide fine-grained feedback. Lightman et al. \cite{lightman2023letsverifystepstep} introduce a human-in-the-loop approach for creating such data via manual annotation, while recent work explores automated methods using Monte Carlo rollouts and tree search to mine supervision signals \cite{luo2024improvemathematicalreasoninglanguage, wang2024mathshepherdverifyreinforcellms, zhang2025lessonsdevelopingprocessreward}. Alternatively, LLM-as-a-judge approaches bypass training by prompting generalist models to assess step correctness directly \cite{madaan2023selfrefineiterativerefinementselffeedback, brown2024largelanguagemonkeysscaling}. While PRMs offer task-specific feedback, LLM-as-a-judge methods provide label-free, generalized supervision \cite{li2024llmsasjudgescomprehensivesurveyllmbased}. Together, these approaches reflect a shift toward scalable, fine-grained guidance for reasoning tasks, paving the way for inference-time strategies that improve performance without retraining.
\\\\
\textbf{Inference-time compute scaling} emerged as a central strategy for improving model output quality under fixed compute budgets \cite{snell2024scalingllmtesttimecompute}. Sampling-based methods generate multiple candidates and apply heuristics to select the best, as seen in best-of-N (BoN) and weighted BoN (WBoN), which sample complete solutions and rerank them using verifiers \cite{brown2024largelanguagemonkeysscaling, huang2025bestofnbestthemcoverage}. Diverse Verifier Tree Search (DVTS) improves efficiency by pruning unpromising paths while preserving diversity \cite{beeching2024scalingtesttimecompute}. Puri et al. recast inference-time scaling as probabilistic inference, resampling partial trajectories based on PRM likelihoods to outperform strong baselines on MATH500 \cite{puri2025probabilisticinferenceapproachinferencetime}. Twisted Sequential Monte Carlo (TSMC) further generalizes this view by sampling from an unnormalized target distribution guided by a terminal potential and learned twist functions \cite{zhao2024probabilisticinferencelanguagemodels}. Search-based strategies elevate partial trajectories into structured forms. Tree-of-Thoughts builds reasoning trees guided by self-evaluation to select promising paths \cite{yao2023treethoughtsdeliberateproblem}, while MCTS-based methods adapt Monte Carlo Tree Search for decoding, using simulations and heuristics to steer generation \cite{zhou2024languageagenttreesearch, guan2025rstarmathsmallllmsmaster, choi2023kctsknowledgeconstrainedtreesearch, zhang2023planninglargelanguagemodels}.
\section{Methodology: Inference-Time Scaling with A*}
\begin{figure}
  \centering
  \includegraphics[width=.99\linewidth]{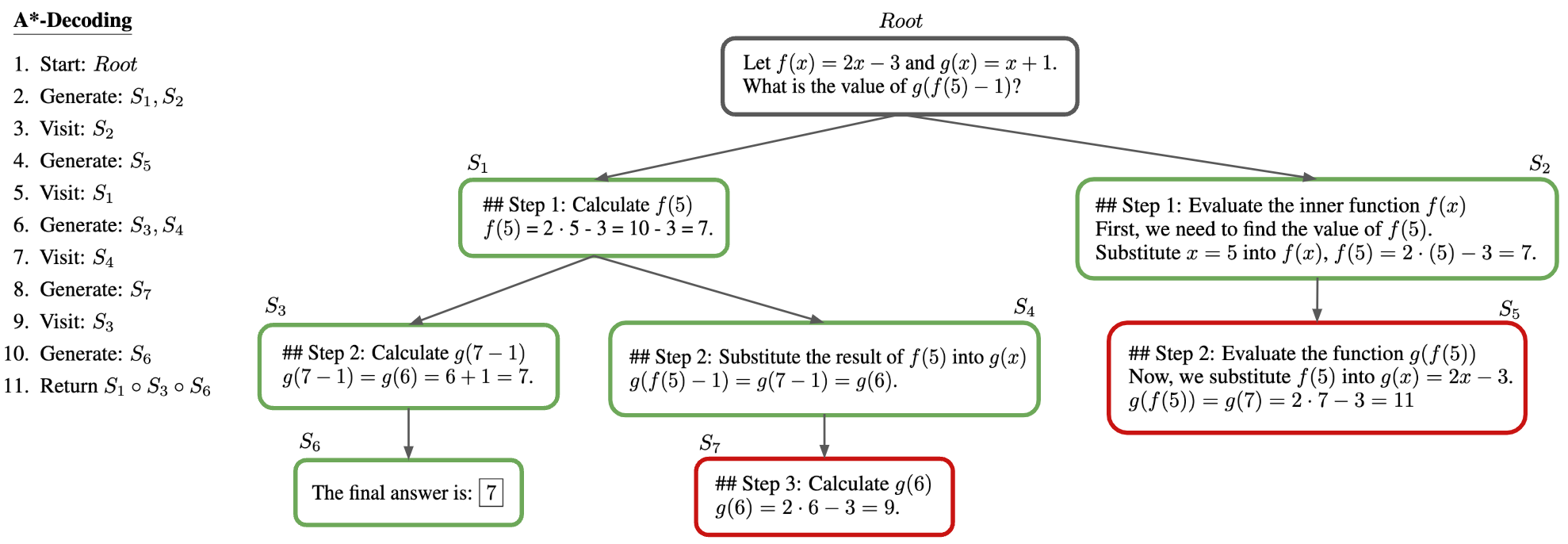}
  \caption{Illustration of A*-decoding applied to a math question involving function composition. At each step, the decoder expands candidate states, computes heuristic values, and selects the most promising path according to the A* total estimated cost. The final solution path, highlighted in green from the lowest green state up to the root, shows how A*-decoding efficiently navigates the solution space to find a correct answer with optimal computation.}
  \label{fig:illustration}
\end{figure}

This work formulates language model inference as state space search, applying the A* algorithm to optimally guide generation toward high-quality completions. A detailed background on state space models and search is provided in Appendix \ref{app:state_space_search_form}. We begin by introducing the A*-algorithm, originally formulated by Hart et al. \cite{4082128} in Section \ref{sec:a_star_search}. We then formalize the state definitions and transition model of the state space model in Section \ref{sec:ssm_llm}. We follow with our A* definitions of the cost and heuristic functions utilizing an external reward signal in the form of a PRM in Section \ref{sec:c_h_def}. Lastly, we go over algorithmic optimizations for efficiently guiding graph exploration under a constrained inference time budget in Section \ref{sec:opt}. We outline the implementation in Algorithm \ref{alg:a_star} and illustrate the execution in Figure ~\ref{fig:illustration}.
\subsection{A* Search}
\label{sec:a_star_search}
A* search is a best-first search algorithm widely used in pathfinding and graph traversal. It builds on Breadth-first search and Dijkstra’s algorithm by incorporating a heuristic that estimates how close a node is to the goal. The total cost of a node is represented as the sum of the cost incurred from the start node and the heuristic estimate to the goal, enabling the algorithm to balance between prioritizing the path taken so far (''exploitation'') and exploring promising paths (''exploration''). At each step, A* selects the node with the lowest total $f(n)$ as defined in Equation \ref{eq:a_star_f}, where $g(n)$ is the cost from the start node to the current node $n$, $h(n)$ is the heuristic estimate of the cost to reach the goal from $n$, and $f(n)$ is the total estimated cost of the cheapest path going through $n$.
\begin{equation}
f(n) = g(n) + h(n)
\label{eq:a_star_f}
\end{equation}
\subsection{Language Model Inference as State Space Search}
\label{sec:ssm_llm}
Given a language model policy $\pi_\theta(x_t \mid x_{<t})$ which defines a distribution over next tokens conditioned on a given context, we can conceptualize the language model's generation process as a traversal through a state space. Let $\mathcal{S}$ be the set of all possible states, $\mathcal{A}$ be the set of all possible actions one can take from a state $s_i \in S$ and $\mathcal{V}$ be the token vocabulary of the policy where $x_i \in \mathcal{V}$. We define a state $s=(x_1,\ldots,x_t)$ to be a collection of tokens forming a partially generated reasoning trajectory effectively segmenting the generated text into discrete, interpretable units of reasoning, where $t < T$ is the token generation limit. The initial state $s_0 \in \mathcal{S}$ corresponds to the input sequence. The set of goal states $\mathcal{G} \subseteq \mathcal{S}$ consists of terminal nodes in the search graph and represent completed solutions. A trajectory is considered to have reached a goal state when it generates an end-of-sequence token within its continuation. In our formulation, each action $a_i \in \mathcal{A}$ corresponds to a proposed continuation of the current partially generated reasoning trajectory. Given a current state $s$, we sample a set of $k$ candidate continuations $C(s) = \{c^{(1)},\ldots,c^{(k)}\}$ from the model policy, each representing a plausible extension of the trajectory. Each of these sampled continuations defines a new successor state $s' = s \circ  c^{(i)}$ where $\circ$ denotes sequence concatenation. The transition model in this framework is inherently hybrid: the generation of successor states is stochastic, governed by the language model’s sampling distribution under different decoding temperatures and randomness controls, while the selection of the next state to visit is deterministic, according to the A* selection process. Specifically, once a set of successor states is generated, A* evaluates each of them using a total score $f(s')=g(s')+h(s')$, where $g$ denotes the accumulated cost of reaching $s'$, and $h$ is a heuristic estimate of the remaining cost (defined in Section \ref{sec:c_h_def}). The next state is selected deterministically as the one with the highest total value (or equivalently, lowest cost under a reward-as-negative-cost formulation). Formally, the transition model can be defined as a two step process:
\begin{enumerate}
    \item \textbf{Sampling step (stochastic)}
    \begin{equation}
    C(s) = \left\{ c^{(1)},\ldots,c^{(k)} \right\} \ \text{where } c^{(i)} \sim \pi(\cdot \mid s)
    \label{eq:sampling}
    \end{equation}

    \item \textbf{Selection step (deterministic)}
    \begin{equation}
    s^* = \arg\min_{s' \in C(s)} f(s') = \arg\min_{s' \in C(s)} \left[ g(s') + h(s') \right]
    \label{eq:selection}
    \end{equation}
\end{enumerate}
A state can be defined at multiple granularity levels depending on the generation task and the heuristic guiding the search. Alternative definitions may involve less granular text segments, such as individual tokens, fixed-size token blocks, or larger units like words and sentences. In some environments such as the game of Countdown or Maze navigation tasks, these coarser segmentations may be more practical and effective for exploration. According to the properties of A* as a single shortest path algorithm, once a trajectory reaches a goal state, it is immediately selected as the final solution without further exploration.
\subsection{A process supervision definition for the A* heuristic}
\label{sec:c_h_def}
\textbf{Heuristic function $h(s)$} \quad In traditional search problems, a heuristic estimates the remaining cost to reach a goal. However, in the context of open-ended generation tasks there is no singular goal state. Instead, the objective is to generate high-quality, correct completions towards a generation task. Due to this nature, we redefine the A* heuristic not as a distance-to-goal estimator, but as a local quality estimate of a candidate continuation aligned with task-specific notions of better outputs. In this work, we propose a process supervision definition to efficiently guide the A* search in the state space. Given a state $s$, we define an external reward function $r(s): S \rightarrow [0,1]$ that assigns a normalized quality score to each partial trajectory. Our heuristic $h(s)$ is then given by:
\begin{equation}
h(s) = 1-r(s)
\label{eq:h_def}
\end{equation}
Consistent with the cost-minimization nature of A*, we invert the heuristic scoring scale and calculate the complement of the reward to ensure lower $h(s)$ values correspond to better partial generations. In our evaluation setup in Section \ref{sec:exp_setup}, we experiment with a PRM as a well-suited candidate definition for $r(s)$, providing the necessary supervision for graph search. Notably, depending on task requirements and inference budget, the reward signal can also come from a shallow verifier (e.g., Pass@1 unit‑test success for code generation) that checks for task‑specific functional correctness, or a LLM-as-Judge (e.g., prompting a large LM to assign ''correct''/''incorrect'' scores) that evaluates candidate continuations using general-purpose knowledge (a potential extension for future research).
\\\\
\textbf{Cost function $g(s)$} \quad Building on our heuristic formulation, we propose a cost function definition that mirrors the structure of traditional distance-based A* search \cite{4082128}. In distance-based A*, cost typically represents a non-negative measure of effort required to transition between states. In our setting, we interpret effort as the degree to which the heuristic value improves, with each reduction signaling meaningful progress through the search space. That is, a decrease $h(s) - h(s')>0$ between successive states reflects movement toward more promising states, analogous to covering positive ground in a spatial graph. We define the function as follows:
\begin{equation}
g(s, s') = \begin{cases} 
h(s) - h(s') & \text{if } h(s') < h(s) \\
0 & \text{otherwise}
\end{cases}
\label{eq:g_def}
\end{equation}
\subsection{Bounding Search Expansion for Tractable Inference}
\label{sec:opt}
In practical scenarios, inference is bounded by limited computational resources, rendering full exhaustive A* search infeasible. To ensure tractable and efficient inference, we introduce a dynamic control mechanism that bounds the number of states explored during inference. Under unconstrained expansion, sampling $k$ continuations per state leads to $O(k^d)$ states at depth $d$. To mitigate this combinatorial explosion, we impose a global cap $b_{max}$ on the number of states that can be added to the graph at each depth level with candidates inserted in order of arrival. Once the threshold $b_{max}$ is reached for a depth level, all remaining candidates at that depth are discarded, regardless of their heuristic value. This strategy bounds the total number of explored nodes to $O(b_{max} \cdot d)$, converting the exponential growth of the search space into a linear function of depth. Section \ref{sec:ablations} presents a detailed ablation.
\begin{algorithm}[t]
\caption{A*-Decoding}
\begin{algorithmic}[1]
\Require prompt $s_0$, LM $\pi_\theta$, heuristic $h$, candidates $k$, 
        scale controls $\!\langle d_{\max},\,b_{\max},\,\tau_h\rangle$
\State $\textit{OPEN}\gets\{(s_0,\,g\!=\!0,\,f\!=\!h(s_0))\}$
\While{$\textit{OPEN}\neq\emptyset$}
    \State $(s,g,f)\gets\textit{OPEN}.\textsc{pop\_min}()$;\quad \textbf{if} $\textsc{EOS}(s)$ \Return $s$
    \State \textbf{if} $\textsc{depth}(s)=d_{\max}$ \Return \textsc{Rollout}$(s)$
    \For{$c\in\textsc{Sample}(\pi_\theta,s,k)$}
        \State $s'\gets s\Vert c$;\; \textbf{if} \textsc{Prune}$(s',b_{max},\tau_h)$ \textbf{continue}
        \State $g'\gets g+\max(0,h(s)-h(s'))$;\; $f'\gets g'+h(s')$
        \State $\textit{OPEN}.\textsc{push}(s',g',f')$
    \EndFor
\EndWhile
\end{algorithmic}
\label{alg:a_star}
\end{algorithm}
\section{Experiments}
In this section, we present the experiments conducted to evaluate the effectiveness of our proposed A* decoding strategy. We begin by outlining our experimental setup, including benchmarks, models, baselines, and evaluation metrics (Section \ref{sec:exp_setup}). We then report our main results, comparing A* decoding to other inference-time scaling methods, as well as to larger open-source models on exact match accuracy (Section \ref{sec:main_results}). Next, we analyze the average number of tokens generated per sample during inference for each scaling method, as a proxy for inference-time token efficiency (Section \ref{sec:token_efficiency}). Finally, we perform extensive ablation studies on key hyperparameters and design choices to understand their impact on both performance and inference efficiency (Section \ref{sec:ablations}).
\subsection{Experimental setup}
\label{sec:exp_setup}
\textbf{Benchmarks} \quad We evaluate our proposed method and baselines on MATH500 and AIME 2024. The datasets are selected to cover a wide range of problem types and difficulty levels, enabling a thorough evaluation of model performance and token efficiency.\\\\
\textbf{Models} \quad We conduct experiments on a diverse set of SLMs chosen to represent a range of parameter scales within their category. We experiment with very compact SLMs such as Llama-3.2-1B-Instruct and Qwen3-1.7B, as well as larger models like Llama-3.1-8B-Instruct and Qwen3-8B \cite{grattafiori2024llama3herdmodels, qwen3}. This range allows us to assess the effectiveness of our approach across architectures and sizes, and to explore how performance scales with stronger/weaker base models under fixed decoding strategies. Inference with the Qwen3 model family is performed with thinking mode disabled \cite{qwen3}.\\\\
\textbf{Process Supervision}\quad Puri et al. \,\cite{puri2025probabilisticinferenceapproachinferencetime} report that, among the PRMs they benchmarked on MATH500 and AIME 2024, Qwen 2.5‑Math‑PRM‑7B exhibited the steepest inference‑time scaling curve. Motivated by that finding, we adopt the same PRM as our primary evaluation signal. Concretely, at each search state we query the PRM once and keep only the \textit{final} reward token emitted (see Appendix \ref{app:prompt_templates}). This final token serves as an aggregated estimate of reward over the partial trajectory and is used directly as our heuristic $h(s)$. Zhang et al.\,\cite{zhang2025lessonsdevelopingprocessreward} independently confirm that reward aggregation is more robust than \textit{min}, \textit{product}, or \textit{last}, providing further justification for our choice. A comprehensive ablation over alternative PRMs appears in Section \ref{sec:ablations}.\\\\
\textbf{Baselines} \quad We compare our approach against a range of strong inference-time reasoning strategies. These baselines represent diverse approaches to structured decoding, sampling, and prompt-based reasoning. Specifically, we evaluate the following methods.
\begin{itemize}
    \item \textbf{Chain-of-Thought}: Serves as an out-of-the-box baseline.
    \item \textbf{Best-of-N}: Selects the highest-scoring answer from N independently sampled completions, where scoring is done using a PRM.
    \item \textbf{Self-Consistency}: Picks the most frequent answer from multiple CoT samples.
    \item \textbf{Particle Filtering}: Generates multiple trajectories by iteratively sampling next steps, with probabilities weighted by the PRM scores of the current partial solutions.
\end{itemize}
\textbf{Evaluation} \quad We evaluate model performance using exact match accuracy and inference budget usage. To ensure consistency, all models are prompted with the same CoT template (see Appendix \ref{app:prompt_templates}), encouraging structured reasoning and comparable outputs. Final answers are normalized into a unified symbolic form, allowing for equivalence across mathematically identical expressions. More details on our scoring procedure are provided in Appendix \ref{app:scoring}.
\subsection{Main results}
\label{sec:main_results}
We summarize our main results in Table \ref{tab:main-results} and outline the key insights below.
\begin{itemize}
    \item Across all baselines strategies, \textbf{A*-decoding delivers the most token-efficient performance}, maximizing the accuracy gained per token generated.
    \item A*-decoding closes the performance gap between small and large models, \textbf{enabling SLMs to exceed the accuracy of much larger counterparts}. Qwen-1.7B with A* surpasses the reasoning accuracy of o1 on MATH500, while Llama-3.1-8B-Instruct reaches the performance of 70x larger Llama-3.1-70B-Instruct-Turbo on AIME 2024.
    \item A*-decoding delivers \textbf{consistent performance improvements across model sizes and architectures}, maintaining high accuracy and token efficiency regardless of base model strength. In contrast, alternative strategies show greater variability, as their effectiveness is more sensitive to the intrinsic quality and diversity of the model’s raw generations (e.g., Qwen3 being a stronger base than Llama3).
\end{itemize}
\begin{table}
  \centering
  \scriptsize
  \sisetup{table-format=2.1}
  \renewcommand{\arraystretch}{1.2}
  \definecolor{astarcolor}{RGB}{217,217,252}
  \setlength{\tabcolsep}{3pt}
  \begin{tabular}{l
                  S[table-format=2.1]S[table-format=2.1]
                  S[table-format=2.1]S[table-format=2.1]}
    \toprule
    \multirow{2}{*}{\textbf{Model}} &
      \multicolumn{2}{c}{\textbf{MATH500}} &
      \multicolumn{2}{c}{\textbf{AIME 2024}} \\
    \cmidrule(lr){2-3}\cmidrule(lr){4-5}
    & {\textbf{Acc (\%)}} & {\textbf{Tokens (↓)}} 
    & {\textbf{Acc (\%)}} & {\textbf{Tokens (↓)}}\\
    \midrule
    Llama-3.1-70B-Instruct-Turbo        & 65.2 & \multicolumn{1}{c}{391}  & 16.6 & \multicolumn{1}{c}{1,038}  \\
    Qwen2.5-72B-Instruct-Turbo        & \text{81.4} & \multicolumn{1}{c}{496}  & 13.3 & \multicolumn{1}{c}{860}  \\
    OpenAI-o1        & 88.2 & \multicolumn{1}{c}{1,276}  & \text{40.0} & \multicolumn{1}{c}{2,021}  \\
    \midrule
    \multicolumn{5}{l}{\textit{Llama-3.2-1B-Instruct}}\\
    \hspace{1em}Pass@1                 & 26.6 & \multicolumn{1}{c}{583}   & \text{0.0} & \multicolumn{1}{c}{737}   \\   
    \hspace{1em}Self-Consistency   & 38.4 & \text{31,043} & 3.3 & \text{52,761} \\
    \hspace{1em}Best-Of-N                & 51.4 & \text{30,783} & 3.3 & \text{52,440} \\
    \hspace{1em}Particle Filtering & 64.0 & \text{38,856} & 10.0 & \text{76,654} \\
    \rowcolor{astarcolor}
    \hspace{1em}\textbf{A* Decoding (ours)}
                                   & \textbf{64.4} & \text{\textbf{35,365}}
                                   & \textbf{13.3} & \text{\textbf{80,108}}\\
    \midrule
    \multicolumn{5}{l}{\textit{Llama-3.1-8B-Instruct}}\\
    \hspace{1em}Pass@1                 & 47.2 & \multicolumn{1}{c}{581}   & 3.3 & \multicolumn{1}{c}{821}   \\   
    \hspace{1em}Self-Consistency   & \text{65.0} & \text{35,357} & 6.6 & \text{64,506} \\
    \hspace{1em}Best-Of-N                & 72.6 & \text{35,150} & \text{10.0} & \text{65,155} \\
    \hspace{1em}Particle Filtering & \textbf{77.0} & \textbf{36,291} & 16.6 & \text{95,403} \\
    \rowcolor{astarcolor}
    \hspace{1em}\textbf{A* Decoding (ours)}
                                   & 76.4 & \text{37,567}
                                   & \textbf{16.6} & \textbf{94,615}\\
    \midrule
    \multicolumn{5}{l}{\textit{Qwen3-1.7B}}\\
    \hspace{1em}Pass@1                 & 68.2 & \multicolumn{1}{c}{617}   & \text{10.0} & \multicolumn{1}{c}{1,532}   \\   
    \hspace{1em}Self-Consistency   & 78.4 & \text{38,469} & 13.3 & \text{94,621} \\
    \hspace{1em}Best-Of-N                & \text{85.2} & \text{38,558} & \text{20.0} & \text{94,086} \\
    \hspace{1em}Particle Filtering & 79.8 & \text{30,994} & 16.6 & \text{55,309} \\
    \rowcolor{astarcolor}
    \hspace{1em}\textbf{A* Decoding (ours)}
                                   & \textbf{83.0} & \textbf{28,307}
                                   & \textbf{16.6} & \textbf{45,139}\\
    \midrule
    \multicolumn{5}{l}{\textit{Qwen3-8B}}\\
    \hspace{1em}Pass@1                 & 79.6 & \multicolumn{1}{c}{527}   & \text{20.0} & \multicolumn{1}{c}{1,346}   \\   
    \hspace{1em}Self-Consistency   & 86.2 & \text{33,388} & 33.3 & \text{89,005} \\
    \hspace{1em}Best-Of-N                & 89.4 & \text{33,344} & 36.6 & \text{87,757} \\
    \hspace{1em}Particle Filtering & \text{87.0} & \text{29,412} & \text{23.3} & \text{61,066} \\
    \rowcolor{astarcolor}
    \hspace{1em}\textbf{A* Decoding (ours)}
                                   & \textbf{88.2} & \textbf{23,037}
                                   & \textbf{30.0} & \textbf{57,748}\\
    \midrule
    \bottomrule
  \end{tabular}
  \vspace{0.6em}
  \caption{Exact match accuracy and respective average total generated tokens per sample on the MATH500 and AIME 2024 benchmarks across different LLMs and inference-time scaling approaches. Results in bold correspond to the highest accuracy over tokens ratio among the four scaling methods, indicating the marginal accuracy gain per generated token. Each baseline strategy is evaluated with a computational budget of $k=64$ model generations while A*-decoding with the proposed optimal setting of $k=16$ candidate continuations.}
  \label{tab:main-results}
\end{table}
\subsection{Scaling inference-time exploration}
\label{sec:token_efficiency}
Figure \ref{fig:cost_ablation} analyzes how scaling inference-time exploration affects token and PRM efficiency, comparing A*-decoding to best-of-N, self-sonsistency, and particle filtering. \textbf{In our results, we find that A*  delivers up to 3x greater overall efficiency than the next-best strategy.} With Llama-3.2-1B-Instruct, best-of-N requires around 32k tokens to surpass 50\% overall accuracy, whereas A* achieves the same performance with just 18k tokens. This efficiency gap persists at the 8B scale, where best-of-N requires about 17k tokens to reach 70\% in accuracy while A* matches it in only 8k tokens. In our experiments, particle filtering follows a similar scaling trajectory, but at a significantly higher PRM cost. A* reduces PRM inference passes by about 30\% by concentrating compute on the most promising derivations rather than resampling full particle sets. This leads to reduced overall latency and cost, particularly in settings where the PRM has a substantially higher parameter count to provide  precise supervision. The compute advantage of A* becomes more pronounced with smaller models, highlighting that strategically allocating inference-time compute can outperform brute-force methods and enable more efficient deployment in resource-constrained settings.
\subsection{Ablations}
\label{sec:ablations}
\begin{figure}[H]
  \centering
  \includegraphics[width=0.99\linewidth]{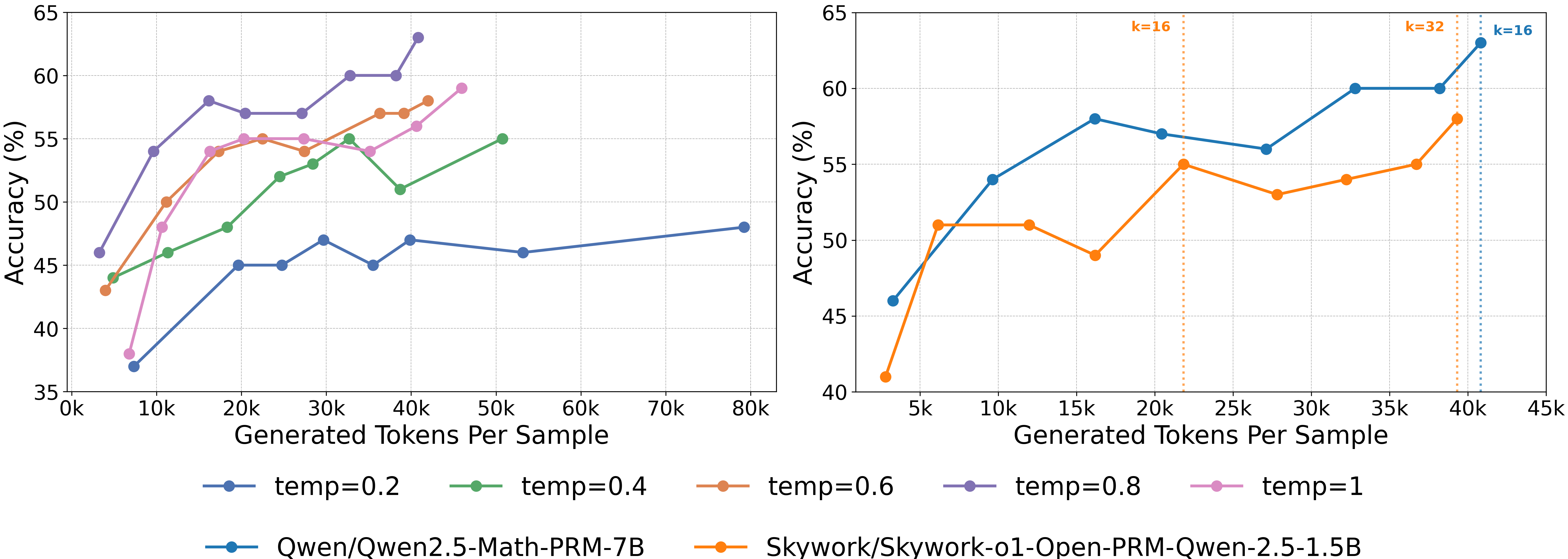}
  \caption{Left: Llama-3.2-1B-Instruct performance on a 100-problem subset of MATH500 with different sampling temperature settings (0.2, 0.4, 0.6, 0.8, 1.0). Right: Llama-3.2-1B-Instruct performance on a 100-problem subset of MATH500 with process supervision signals from different models (Qwen2.5-Math-PRM-7B, Skywork-o1-Open-PRM-Qwen-2.5-1.5B).}
  \label{fig:temp_prm_ablations}
\end{figure}

\textbf{Explore vs exploit trade-off in sampling} \quad We perform an ablation study on decoding sampling temperatures using a subset of 100 problems from the MATH500 benchmark. As shown in Figure \ref{fig:temp_prm_ablations}, we find that a temperature setting of 0.8 offers the best performance, striking a balance between candidate diversity and answer quality. This setting consistently achieves higher final accuracy and optimal token efficiency under a fixed budget, while also demonstrating more stable and reliable scalability. Based on this result, we use temperature 0.8 for all experiments in this work. Notably, this finding is consistent with observations reported by Puri et al. and Beeching et al. \cite{puri2025probabilisticinferenceapproachinferencetime, beeching2024scalingtesttimecompute} who report similar results.\\\\
\textbf{Impact of different process supervision signals} \quad  In Figure \ref{fig:temp_prm_ablations}, we assess the influence of process supervision signal quality on A*-decoding by evaluating two PRMs at different parameter scales: Qwen2.5-Math-PRM-7B and Skywork-o1-Open-PRM-Qwen-2.5-1.5B \cite{zhang2025lessonsdevelopingprocessreward, skyworkopeno12024}. We compare against the smaller Skywork 1.5B model to explore whether small supervision models can guide inference effectively under comparable generation settings. Across experiments on a subset of 100 questions from the MATH500 benchmark, we observe that the 7B PRM significantly outperforms its 1.5B counterpart at all token budgets. Interestingly, the smaller PRM displays lower average token usage despite identical candidate continuation settings, suggesting it fails to sustain in-depth exploration of promising paths. Even as candidate diversity increases up to k=32, the 1.5B PRM does not yield meaningful improvements in accuracy, highlighting its limited ability to reliably distinguish high-quality reasoning trajectories. These observations emphasize the importance of a strong and expressive PRM capable of assigning informative signals across diverse intermediate states.
\textbf{Controlling the graph breadth} \quad  We study the effect of varying the graph breadth parameter $b_{max}$ on performance and token efficiency using Llama-3.2-1B-Instruct evaluated on the MATH500 dataset. As shown in Appendix \ref{app:breadth_ablation}, we compare $b_{max} \in \{5,10,20\}$ and find that smaller values of $b_{max}$ lead to more efficient scaling and better final accuracy. Specifically, $b_{max}=5$ consistently outperforms larger values, reaching close to 65\% accuracy with fewer tokens per sample. Our hypothesis is that lower breadth constraints encourage deeper exploration within a more focused subset of candidates. Limiting the number of candidates retained at each level allows us to sample more from the policy across time steps, increasing the diversity of thought trajectories and the chance of uncovering high-quality reasoning paths. On the contrary, increasing $b_{max}$ leads to diminishing returns with the additional overhead resulting in fewer overall expansions before the compute budget is exhausted. All experiments were conducted with a setting of $b_{max}=5$.
\section{Discussion}
 \textbf{Broader Impact} \quad  A*-decoding inherits biases from the base language model and the external supervision signals used during inference. If either source reflects flawed or biased patterns, the search may reinforce and amplify undesirable reasoning trajectories. \\\\
 \textbf{Limitations} \quad While A*-decoding improves token efficiency and accuracy, it relies heavily on the quality of external supervision. Misaligned heuristics can degrade search performance, especially in open-ended tasks with weak reward signals. Our method also introduces tunable hyperparameters (e.g., sampling temperature, graph breadth, candidate count) that requires tuning to model scale and task.\\\\
 \textbf{Conclusion} \quad A*-decoding augments the ''System 1'' fluency of language models with a ''System 2'' search over reasoning paths, bringing classical ideas from planning into modern inference. This structure enables small models to rival much larger ones under tight compute, while enabling more interpretable reasoning. We see this synthesis of symbolic search and neural generation as a promising direction for scalable model intelligence.

\newpage

\bibliographystyle{plain}
\bibliography{references}

\newpage


\appendix

\section{Appendix}
\subsection{Language Model Inference as State Space Search (Extended)}
\label{app:state_space_search_form}
\textbf{State space models} formalize language models as stochastic processes that sequentially generate text by traversing a discrete space of token sequences. Formally, at time step $t$, the state is the partial sequence $s_t = (x_1,\ldots,x_t)$ where $x_i \in \mathcal{V}$ and $\mathcal{V}$ denotes the vocabulary. The action space is $\mathcal{V}$, with action $a_t \in \mathcal{V}$ corresponding to the selection of the next token. The transition function is deterministic, given by $\mathcal{T}(s_t, a_t)=(x_1,\ldots,x_t,a_t)$, which appends the selected token to the sequence. The language model specifies a stochastic policy $\pi(a_t \mid s_t)=p(a_t \mid s_t)$, defining a probability distribution over actions conditioned on the current state. Consequently, text generation can be viewed as a stochastic traversal of an exponentially large tree of sequences, linking language model inference to classical frameworks of sequential decision-making and search over structured spaces.\\\\
\textbf{State space search} explores possible sequences by navigating the state space according to a defined policy, balancing exploration of novel states and exploitation of high-probability transitions. The objective is to find sequences that optimize a heuristic, which may encode factors such as task objectives or structural properties. Formally, the search problem is defined by the tuple $(\mathcal{S},\mathcal{A},\mathcal{T},s_0,\mathcal{G})$, where $\mathcal{S}$ denotes the set of partial sequences (states), $\mathcal{A}$ the set of available token actions, $\mathcal{T}: \mathcal{S} \times \mathcal{A} \rightarrow \mathcal{S}$ the transition function mapping a state-action pair to a successor state, $s_0 \in \mathcal{S}$ the initial prompt, and $\mathcal{G} \subseteq \mathcal{S}$ the set of goal states. Approaching LM inference as state space search motivates the need for search methods that can efficiently identify promising sequences, given the high-dimensional nature of the space of all possible token sequences making exhaustive search intractable.
The A* algorithm introduces several strong theoretical constraints that guarantee completeness and optimality. To validate the applicability of our inference-time scaling approach with the proposed cost and heuristic definitions, we review these theoretical constraints in detail.
\subsection{Theoretical Constraints: Admissibility, Consistency, and Informativeness}
Hart et al. \cite{4082128} introduce a number of theoretical constraints that govern the optimality and efficiency of A* search.
\begin{itemize}[left=0pt]
  \item \textbf{Admissibility} requires that the heuristic never overestimates the true cost to the goal. While this ensures optimality in traditional search, its direct application to language generation is less clear-cut where “optimality” is more ambiguous. In our setting, we allow for heuristics that may trade strict admissibility for greater informativeness and efficiency, aiming to find high-quality solutions within practical inference limits.
  \item \textbf{Consistency} ensures that the estimated cost to the goal does not decrease after a transition: $h(s) \leq c(s,s') + h(s')$. This constraint prevents re-expansion of nodes and guarantees both correctness and efficiency. Our cost function $c(s,s')=max(0,h(s)-h(s'))$ preserves this condition by enforcing non-negative and monotonic transitions across the search graph.
  \item \textbf{Informativeness} captures how well the heuristic differentiates between better and worse states. A trivial heuristic like $h(s)=0$ satisfies admissibility but collapses A* into uniform-cost search, losing its guidance power. Our reward-based heuristic leverages learned external process supervision signals to prioritize promising candidates and significantly improve search efficiency.
\end{itemize}
\subsection{Parsing and scoring responses}
\label{app:scoring}
To ensure fair and consistent evaluation across mathematically equivalent outputs, we employed a symbolic equivalence check using the math-verify library \cite{Kydlicek_Math-Verify_Math_Verification} in conjunction with sympy, similar to \cite{puri2025probabilisticinferenceapproachinferencetime,beeching2024scalingtesttimecompute}. During evaluation, each model-generated final answer was first extracted and parsed into a structured symbolic expression. The parsing step normalized notational variations (e.g., converting "1/2" and "0.5" into canonical forms), and handled common formatting inconsistencies found in the outputs. The final accuracy as compute as the ratio of exact matches over the total datapoints of evaluation.
\subsection{Breadth $b_{max}$ ablation graph}
\begin{figure}[H]
\label{app:breadth_ablation}
  \centering
  \includegraphics[width=0.8\linewidth]{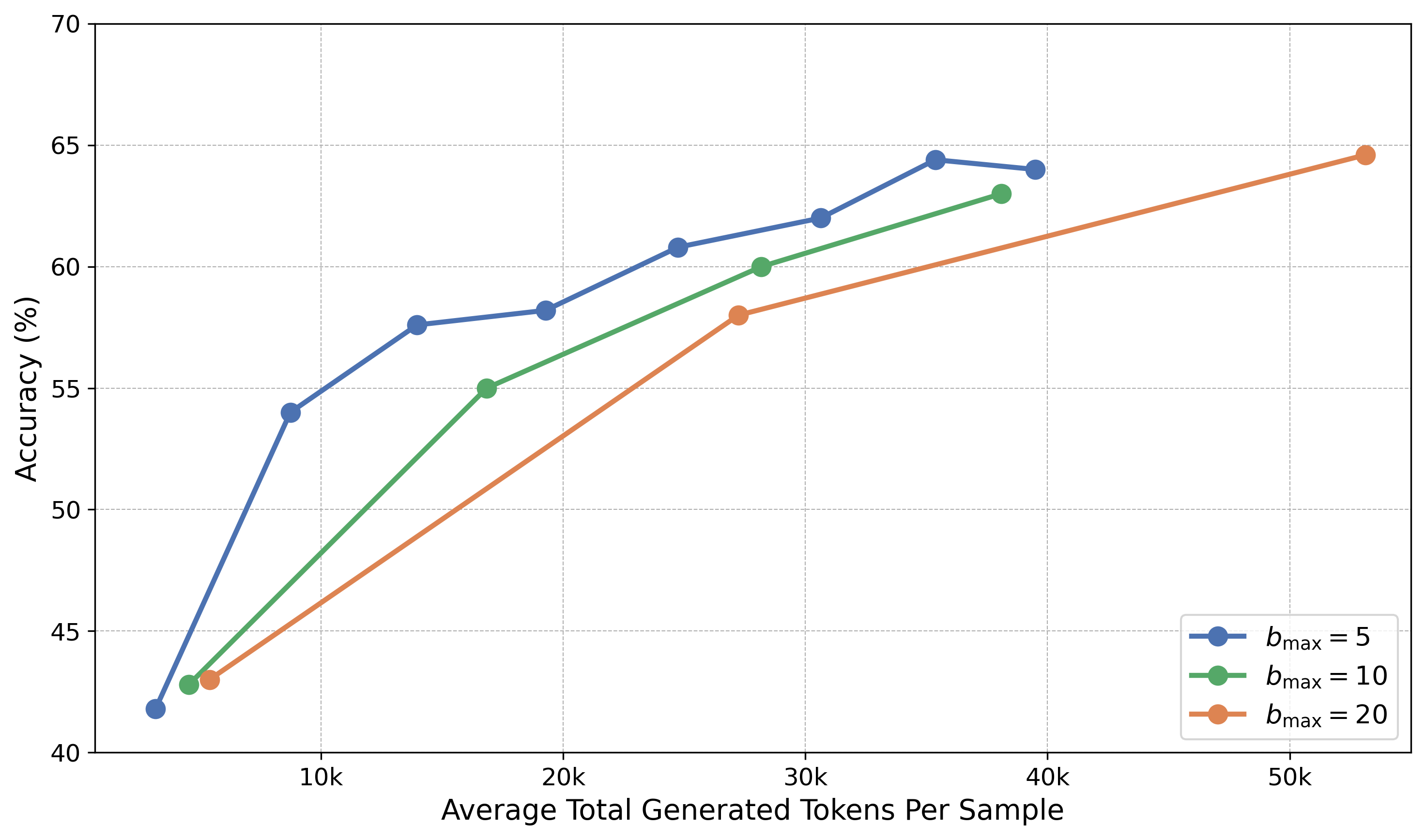}
  \caption{Llama-3.2-1B-Instruct performance on MATH500 with different max breadth $b_{max}$ settings (5, 10, 20). Setiting $b_{max} = 5$ achieves the highest token-efficiency and overall accuracy compared to other settings.}
\end{figure}
As a breadth-limiting mechanism, $b_{max}$ provides a tunable trade-off between computational cost and search completeness, allowing adaptive control over the inference process without requiring full traversal of the generation graph.
\subsection{Inference Prompt Templates}
\begin{tcolorbox}[
    title=\textbf{PRM Evaluation Instructions Aggregate Scoring},
    coltitle=black,
    colframe=black,
    colback=white,
    colbacktitle=gray!20,
    sharp corners,
    fonttitle=\bfseries,
    boxrule=0.5pt,
    left=2mm,
    right=2mm,
    top=1mm,
    bottom=1mm,
    enhanced,
]
\#\# Step 1: \text{[Concise description]}\\
\text{[Brief explanation and calculations]}\\\\
\#\# Step 2: \text{[Concise description]}\\
\text{[Brief explanation and calculations]}\\\\
\ldots
\\\\
<aggregate\_reward>
\end{tcolorbox}
\label{app:prompt_templates}
\begin{tcolorbox}[
    title=\textbf{LLM Chain-Of-Thought Evaluation Instructions},
    coltitle=black,
    colframe=black,
    colback=white,
    colbacktitle=gray!20,
    sharp corners,
    fonttitle=\bfseries,
    boxrule=0.5pt,
    left=2mm,
    right=2mm,
    top=1mm,
    bottom=1mm,
    enhanced,
]
Solve the following math problem efficiently and clearly:\\\\
- For simple problems (2 steps or fewer):\\
Provide a concise solution with minimal explanation.\\\\
- For complex problems (3 steps or more):\\
Use this step-by-step format:\\\\
\#\# Step 1: \text{[Concise description]}\\
\text{[Brief explanation and calculations]}\\\\
\#\# Step 2: \text{[Concise description]}\\
\text{[Brief explanation and calculations]}\\\\
\ldots
\\\\
Regardless of the approach, always conclude with:\\\\
Therefore, the final answer is: boxed\{answer\}. I hope it is correct.\\\\
Where \text{[answer]} is just the final number or expression that solves the problem.
\end{tcolorbox}
\subsection{Experiments compute resources}
All inference experiments were conducted on two A100 GPUs (40GB each). The most compute-intensive runs on the MATH500 benchmark, configured with high values for the number of candidate continuations $k$ and the breadth parameter $b_{max}$ to encourage in-depth exploration, required between 16 to 24 hours to complete. Under recommended hyperparameter settings, typical runs on MATH500 completed within a few hours to a maximum of approximately 10 hours, depending on the size of the base model.

\end{document}